%% file: mem.tex
\documentclass[10pt,twocolumn,letterpaper]{article}

\usepackage{cvpr}
\usepackage{times}
\usepackage{epsfig}
\usepackage{graphicx}
\usepackage{amsmath}
\usepackage{amssymb}
\usepackage{bm}
\usepackage{tabularx}
\usepackage[scientific-notation=true]{siunitx}

\usepackage[pagebackref=true,breaklinks=true,letterpaper=true,colorlinks,bookmarks=false]{hyperref}

\cvprfinalcopy 

\newcommand*\samethanks[1][\value{footnote}]{\footnotemark[#1]}

\ifcvprfinal\fi
\begin{document}

\title{Few-Shot Object Recognition from Machine-Labeled Web Images}

\author{Zhongwen Xu\thanks{Indicates equal contribution.} \hspace{2em}Linchao Zhu\samethanks \hspace{2em}Yi Yang\\
	University of Technology Sydney\\
	{\tt\small \{zhongwen.s.xu, zhulinchao7, yee.i.yang\}@gmail.com}
}

\maketitle

\input{abstract.tex}
\input{intro.tex}
\input{related_work.tex}
\input{model.tex}
\input{expr.tex}
\input{conclusion.tex}

{\small
\bibliographystyle{ieee}
\bibliography{mem}
}

\end{document}

%% file: abstract.tex

\begin{abstract}
With the tremendous advances of Convolutional Neural Networks (ConvNets) on object recognition, we can now obtain reliable enough machine-labeled annotations easily by predictions from off-the-shelf ConvNets. In this work, we present an ``abstraction memory'' based framework for few-shot learning, building upon machine-labeled image annotations.
Our method takes some large-scale machine-annotated datasets (\eg, OpenImages) as an external memory bank. 
In the external memory bank, the information is stored in the memory slots with the form of key-value, where image feature is regarded as key and label embedding serves as value.
When queried by the few-shot examples, our model selects visually similar data from the external memory bank, and writes the useful information obtained from related external data into another memory bank, \ie~{abstraction memory}.
Long Short-Term Memory (LSTM) controllers and attention mechanisms are utilized to guarantee the data written to the abstraction memory is correlated to the query example.
The abstraction memory concentrates information from the external memory bank, so that it makes the few-shot recognition effective.
In the experiments, we firstly confirm that our model can learn to conduct few-shot object recognition on clean human-labeled data from ImageNet dataset. Then, we demonstrate that with our model, machine-labeled image annotations are very effective and abundant resources to perform object recognition on novel categories. Experimental results show that our proposed model with machine-labeled annotations achieves great performance, only with a gap of 1\% between of the one with human-labeled annotations.
\end{abstract}

%% file: intro.tex

\section{Introduction}
\begin{figure}[t]
\centering
\includegraphics[width=0.75\linewidth]{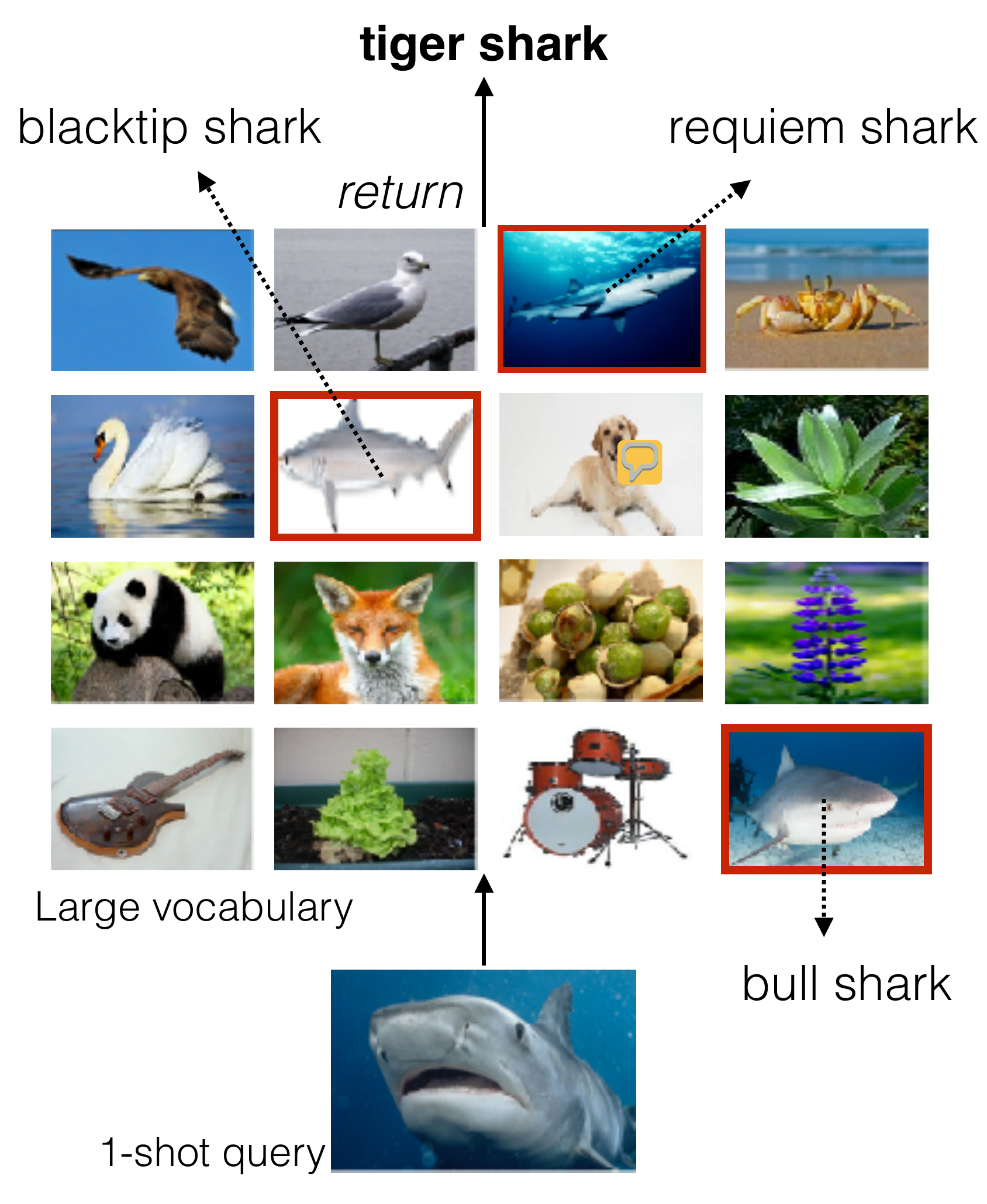}
\caption{Given a large vocabulary of labels and their corresponding images, we would like to conduct few-shot learning on a novel category, which is not in the vocabulary and only have a handful of positive examples. The image examples in the vocabulary are stored in the external memory of our model. The image example from the novel category comes and queries the external memory. Our model reads out helpful information according to visual similarity and LSTM controllers. The retrieved information, \ie, visual features and their corresponding labels are combined to classify this query image example.}
\label{fig:overview}
\end{figure}

Driven by the innovations in the architectures of Convolutional Neural Networks (ConvNets)~\cite{LeNet,GoogLeNet,VGGNet,ResNet}, tremendous improvements on image classification have been witnessed in the past few years. 
With the increase in the capacity of neural networks, the demand on more labeled data of richer categories is rising.
However, it is unlikely and very expensive to manually label a dataset 10 times larger than ImageNet.
It is time to design a new paradigm to utilize the machine-labeled image annotations and enable rapid learning from novel object categories.  Figure~\ref{fig:overview} shows an illustration of the proposed task. Here comes our major question in this work: Given the \emph{machine-labeled} web image annotations, can we conduct object recognition \emph{rapidly} for \emph{novel categories} with \emph{only a handful of examples}?

We propose a new memory component in neural networks, namely \emph{abstraction memory}, to concentrate information from the external memory bank, \eg, large-scale object recognition datasets like ImageNet~\cite{ImageNet} and OpenImages~\cite{openimages}, based on few-shot image queries.
Previous methods which try to learn among different categories or different datasets usually use a larger dataset for pre-training and then conduct fine-tuning on a relatively small dataset. 
The information of the large datasets is encoded in the learnable weights of the neural networks.
Different from previous works, our model utilizes content-based addressing mechanism with a Long Short-Term Memory (LSTM) controller to decide where to read from and where to write into the memories automatically. Given the query image, the neural network applies soft attention mechanism~\cite{soft_att} to find the appropriate information to readout from the external memory and write into another memory. The abstraction memory records helpful information for the specific few-shot object recognition, so that the classification network can utilize readouts from the abstraction memory to recognize the objects from novel categories.

Comparing with previous methods which only discover the relationship of the word embeddings~\cite{jozefowicz2016exploring,word2vec} among the category labels,
we fully utilize visual similarity between the examples of few-shot categories and the external memory bank to make the proposed framework more robust to noisy labels.
If the data of external memory is inconsistent with its label, this sample will be rejected during the visual matching process.
This property make the usage of large-scale machine annotated dataset, \eg, OpenImages~\cite{openimages} feasible.
The machine-labeled annotations for images could be predicted by off-the-shelf ConvNet models (\eg, ResNets~\cite{ResNet}). These annotations are reasonably good but imperfect. 
In this scenario, external dataset can also be images obtained by querying keywords in search engines (\eg, Google Images), and images crawled from social photos sharing sites (\eg, Flickr).
In the experiment section, we show that our proposed method differs machine-annotated data with human-labeled data in a minor gap $\approx 1\%$.

When the novel categories arrive, the network would query and access the external memory, retrieve the related information, and then write differentially into abstraction memory. 
We organize the memories in the data structure \verb|key|:\verb|value|, which was firstly proposed in Key-Value Memory Networks (KV-MemNNs)~\cite{KV-MemNN}. We note that we have a very different implementation from the KV-MemNNs in our model, including LSTM controllers, abstraction memory, and reading mechanisms. Moreover, KV-MemNNs were developed under natural language understanding, and their memory accesses are limited to most recent a few sentences. We extend the key-value storage concept into computer vision applications by novel modifications to enable scalability.
We formulate the image embedding as the \verb|key| and the word embedding of the annotated label as the \verb|value|. 
The additional memory for abstraction extracts information from the external memory and learn task-specific representation for the few-shot learning while maintaining the efficiency.

Our contributions are as follows. 
\begin{enumerate}
	\item We propose a novel task to learn few-shot object recognition upon machine-labeled image annotations. We demonstrate that with the reliable enough machine-labeled annotations, it can achieve great performance with a minor gap (about 1\% accuracy) compared to learning from human-labeled annotations;
	\item  We propose a novel memory component, namely abstraction memory, into the Memory Networks~\cite{MemNN} structure. The abstraction memory can  alleviate the time-consuming content-based addressing of the external memory, enabling the model to be scalable and efficient;

	\item  We utilize both visual embeddings and label embeddings in a form of key-value to make the system robust to the imperfect labeling.
	Hence it can learn from the machine-labeled web images to obtain rich signals for visual representation, which is very suitable for real-world vision application.
	Specifically, we conduct few-shot learning of unseen visual categories, making rapid and accurate predictions without extensive iterations of positive examples.
\end{enumerate}
We demonstrate advantages over state-of-the-art models such as Matching Networks~\cite{vinyals2016matching}, KV-MemNNs~\cite{KV-MemNN}, Exemplar-SVMs~\cite{exemplar}, and Nearest Neighbors~\cite{PRML} on few-shot object recognition tasks.

%% file: related_work.tex

\section{Related Work}

\noindent \textbf{Learning Visual Features from the Web.} Chen~\etal~\cite{NEIL} propose a never ending image learner (NEIL) to extract common sense relationships and predict instance-level labels on web images. NEIL bootstraps the image classifier by training from top-ranked images in Google images as positive samples, then a semi-supervised learning method is used to mine object relationships. Divvala~\etal~\cite{divvala2014learning} leverage Google Books to enrich the visual categories into very broad ranges, including actions, interactions, and attributes. These works focus on mining relationships between objects and intra-class, however, these approaches are prone to errors, since the classification mistakes would accumulate along the iteration procedure due to the bootstrapping nature. Joulin~\etal~\cite{joulin2015learning} argue that ConvNets can learn from scratch in a weakly-supervised way, by utilizing 100M Flickr images annotated with noisy captions. Our work utilizes established state-of-the-art human-level ConvNets to alleviate the error that could come from seed images. We focus on a different task of learning few-shot classification rapidly by benefiting from the rich vocabulary of the web resources.

\noindent \textbf{External Memory in Neural Networks.}  Neural Turing Machines (NTMs)~\cite{NTM} and Memory networks (MemNNs)~\cite{MemNN} are two recently proposed families of neural networks augmented with external memory structure. NTMs are fully differentiable attempts of  Turing machines neural network implementation which learn to read from and write into the external memory. NTMs were demonstrated success on tasks of learning simple algorithms such as copying input strings and reversing input strings. MemNN was proposed to reason from facts/story for question answering, building the relationships among ``story'', ``question'' and ``answer''. End-to-end Memory Networks (MemN2N)~\cite{MemN2N} eliminate the requirements of strong supervision of MemNNs and train the networks in an end-to-end fashion. Key-Value Memory Networks (KV-MemNNs)~\cite{KV-MemNN} incorporate structural information in the form of key-value which makes more flexibility ways to store knowledge bases or documents. Though yielding excellent performance on toy question-answering benchmarks, Memory Networks applications are still limited in natural language understanding domains. We realize the great expressive power of neural networks augmented with external memory, and build upon these great works to learn rapid visual classification from machine-labeled images.

\noindent \textbf{One-shot Learning.} Training neural networks notoriously require thousands of examples for each category, which means conventional neural models are highly data inefficient. Fei-Fei~\etal~\cite{fei2006one} pioneered one-shot learning of object categories and provided an important insight: taking advantages of knowledge learned from previous categories, it is possible to learn about a category from just one, or a handful of images~\cite{fei2006one}. Inspired by the Bayesian Program Learning (BPL) for concept abstraction in Lake~\etal~\cite{lake2015human} and augmented memory neural structures~\cite{NTM,MemNN}, Memory-Augmented Neural Networks (MANNs)~\cite{santoro2016one} utilize meta-learning paradigm to learn the binding of samples and labels from shuffled training batches. Matching networks~\cite{vinyals2016matching} employ metric learning and improve over MANNs significantly by utilizing the attention kernel and the set-to-set framework~\cite{set2set}.

%% file: model.tex

\section{Proposed Approach}

\subsection{Preliminaries}

We briefly introduce some technical preliminaries, including visual features, label embedding and Memory Network variants before going into our proposed model.

\subsubsection{Visual Feature}
We forward the image $I$ through an ImageNet pre-trained ConvNet model, \ie, ResNet~\cite{ResNet}, to extract the visual features for the image. 
The feature extraction procedure is as following,
\begin{equation}
	\bm{x} = \Phi_\text{img}(I),
	\label{eq:visual feature}
\end{equation}
where $\Phi_\text{img}$ is ImageNet pre-trained ResNet-200 model with the final classification layer removed, 
$\bm{x}$ is the visual feature (or embedding) of the image $I$, with 2,048 dimensions.

\subsubsection{Label Embedding}
Instead of one-hot representation of label information, Frome~\etal~\cite{frome2013devise} shows that incorporation of semantic knowledge helps the visual neural models learn to generalize across object categories. 
We denote the label embedding as,
\begin{equation}
	\bm{y} = \Phi_{\text{label}}(l),
	\label{eq:label_feature}
\end{equation}
where $\Phi_{\text{label}}$ is a pre-trained word embedding model, $l$ is the object label, 
and $\bm{y}$ is the vector representation of $l$. 
In this work, we use the state-of-the-art language model~\cite{jozefowicz2016exploring} pre-trained on One Billion Word Benchmark as word embeddings. 
The label embedding dimension is of 1,024.

\subsubsection{Memory Networks} 
Memory Networks (MemNNs)~\cite{MemNN} are a new family of learning models which augment the neural networks with \emph{external memory}. 
The major innovation of Memory Networks is the long-term memory component $\bm{\mathcal{M}}$, which enables the neural networks to reason and access the information from a long-term storage. 
 End-to-End Memory Networks (MemN2N)~\cite{MemN2N} implement Memory Networks in a continuous form, so that end-to-end training becomes feasible. 
The recently proposed Key-Value Memory Networks (KV-MemNNs)~\cite{KV-MemNN} extend MemNNs~\cite{MemNN} and MemN2N~\cite{MemN2N} with structural information storage in the memory slots. 
Instead of having only single vector representation in the memory component as in MemN2N, KV-MemNNs make use of pairs of vectors in the memory slots, \ie, \verb|key|: \verb|value|. 
The incorporation of structural storage of Key-Value form into the memory slots brings a lot more flexibility, which enriches the expressive power of the neural networks. 
The Key-Value property makes information retrieval from the external memory become natural. 

The Memory Network variants (MemNNs, MemN2N, and KV-MemNNs) have been proposed for natural language understanding, where researchers often only validate these models on question answering tasks like bAbI tasks~\cite{weston2015towards}.

\begin{figure*}[t]
\centering
\includegraphics[width=0.75\linewidth]{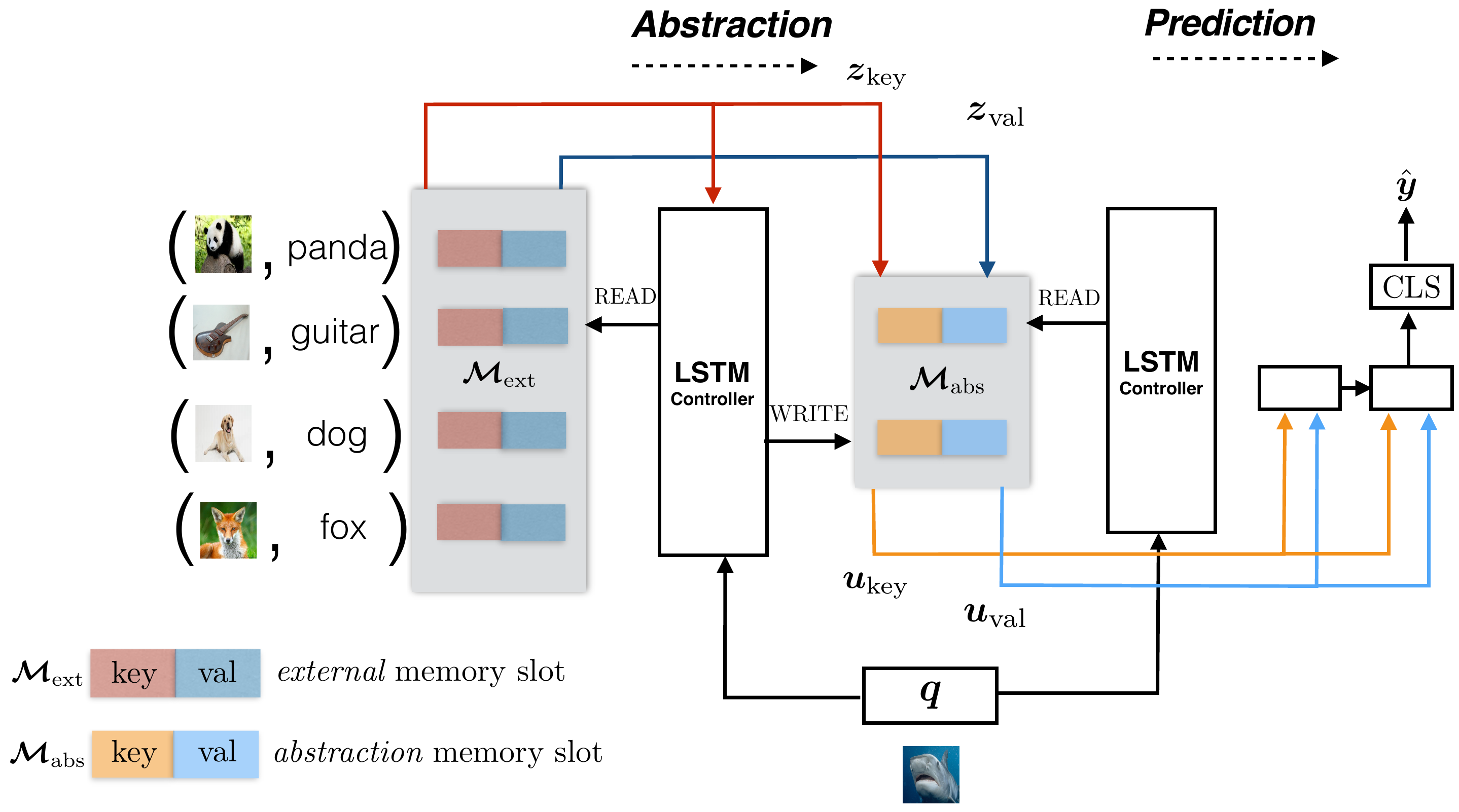}
\caption{An illustration of our proposed model. Best viewed in color.}
\label{fig:framework}
\end{figure*}

\subsection{Model Overview}

In this work, we propose a novel architecture of Memory Networks to tackle the few-shot visual object recognition problem. It keeps the key-value structure, but different from KV-MemNNs, we utilize Long Short-Term Memory (LSTMs) as a ``controller'' when accessing and writing to the memories. Moreover, we introduce a novel memory component, namely \emph{abstraction memory}, to enable task-specific feature learning and obtain scalability. The distinct nature of our proposed abstraction memory makes the neural network ``remember'' the ever presented external memories, analogy to the memory cell $\bm{c}$ in LSTMs but much more expressive. The incorporation of abstraction memory enables stochastic external memory training, \ie, we can sample batches from the a huge external memory pool. Contrast to our work, existing Memory Networks limit their access to external memory to a very small number, \eg, MemN2N limit their access to external memory to most recent 50 sentences~\cite{MemN2N}.

The overview of our model is shown in Figure~\ref{fig:framework}. The whole procedure of our proposed model is illustrated as follows, we re-formulate \verb|key|: \verb|value| as (key, value) in the rest of this work.
\begin{eqnarray}
\bm{q}, \bm{\mathcal{M}}_\text{ext} & = & \verb|EMBED|( I, \{\mathcal{I}_\text{web}, \mathcal{L}_\text{web} \} )   \\
(\bm{z}_\text{key}, \bm{z}_\text{val}) & = & \verb|READ| (\bm{q}, \bm{\mathcal{M}}_\text{ext}), \label{eq:read_ext}\\
\bm{\mathcal{M}_\text{abs}} & \leftarrow & \verb|WRITE| (\bm{q}, (\bm{z}_\text{key}, \bm{z}_\text{val}), \bm{\mathcal{M}}_\text{abs}), \label{eq:write_abs}\\
(\bm{u}_\text{key}, \bm{u}_\text{val}) & = & \verb|READ| (\bm{q}, \bm{\mathcal{M}}_\text{abs}), \label{eq:read_abs} \\
\bm{\hat{y}} & = & \verb|CLS|([\bm{u}_\text{key}, \bm{u}_\text{val}]).  \label{eq:cls}
\end{eqnarray}
\noindent We elaborate each of the operation in the procedure, all of the following operations are parameterized by neural networks:

\begin{enumerate}
	\item \verb|Embed| is a transformation from  the raw inputs to their feature representation, as mentioned in Eqn.~(\ref{eq:visual feature}) and Eqn.~(\ref{eq:label_feature}). Given an image $I$ from a novel category, and a bunch of web images with labels, denoted as $\mathcal{I}_\text{web}$ and $\mathcal{L}_\text{web}$, where $\mathcal{I}$ is the image set and $\mathcal{L}$ is the label set.  Here, input image $I$ is sampled from unseen categories, and the embedded feature for the query image is referred to as query $\bm{q}$ following the notation in Memory Networks. The web images are embedded into the external memory $\bm{\mathcal{M}}_\text{ext}$ through the same embedding networks $\Phi_\text{img}$ and $\Phi_\text{label}$;
	\item \verb|READ| takes the query $\bm{q}$ as input, conducts content-based addressing on the external memory $\bm{\mathcal{M}}_{\text{ext}}$, to find related information according to similarity metric with $\bm{q}$. 
	The external memory is also called the support set in Memory Networks.
	The output of the \verb|READ| is a pair of vectors in key-value form, \ie, $(\bm{z}_\text{key}, \bm{z}_\text{val})$, as shown in Eqn.~(\ref{eq:read_ext});

	\item \verb|WRITE| takes a query $\bm{q}$, key-value pair $(\bm{z}_\text{key}, \bm{z}_\text{val})$ as inputs to 
	conduct write operation.
	The content-based addressing is based on matching input with $\bm{\mathcal{M}_{\text{abs}}}$
	, and then update the content of the corresponding abstraction memory slots as done in Eqn.~(\ref{eq:write_abs});

	\item \verb|READ| from abstraction memory (Eqn.~(\ref{eq:read_abs})) is for the classification stage.
	Take the input query $\bm{q}$ to match with the abstraction memory $\bm{\mathcal{M}}_{\text{abs}}$.
 	Then the obtained pairs of vectors (\ie, ($\bm{u}_\text{key} , \bm{u}_\text{val}$)) are concatenated to be fed into the classification network;

	\item \verb|CLS| operation takes the readout key-value $(\bm{z}_\text{key}, \bm{z}_\text{val})$, concatenates them into one vector $\bm{z}_{\text{cls}} = [\bm{z}_\text{key}, \bm{z}_\text{val}]$. 
	Then $\bm{z}_{\text{cls}}$ goes through a Fully-Connected (\verb|FC|) layer where:
\verb|FC|$(\bm{x}) = \bm{w}^\top \bm{x} + \bm{b}$, and a \verb|Softmax| layer as follows,
\begin{eqnarray}
  \verb|Softmax| (e_i) &= & \frac{\exp(e_{i})}{\sum_{j} \exp(e_{j})} \label{eq:softmax}.
\end{eqnarray}
Section~\ref{sec:cls} shows an LSTM variant of the \verb|CLS| operation.
\end{enumerate}

\subsection{Model Components} \label{sec:model_components}

\subsubsection{Long Short-Term Memory}
In our model, Long Short-Term Memory (LSTMs)~\cite{LSTM} play an important role in the \verb|READ|, \verb|WRITE| and \verb|CLS| procedures and serve as the \emph{controller} of the memory addressing.  LSTMs are a special form of Recurrent Neural Networks (RNNs). LSTMs address the vanishing gradient problem~\cite{bengio1994learning} of RNNs by introducing an \emph{internal} memory cell to encode information from the previous steps. LSTMs have resurged due to the success of sequence to sequence modeling~\cite{seq2seq} on machine translation~\cite{soft_att}, image captioning~\cite{vinyals2015show,karpathy2015deep,xu2015show}, video classification~\cite{yue2015beyond}, video captioning~\cite{venugopalan2015sequence,pan2015hierarchical}, \etc.
Following the notations of Zaremba~\etal~\cite{zaremba2014recurrent} and Xu~\etal~\cite{xu2015show} and assuming $\bm{x}_t \in \mathbb{R}^{D}$, $\bm{T}_{D+d,4d}: \mathbb{R}^{D + d} \rightarrow \mathbb{R}^{4d}$ denotes an affine transformation from $\mathbb{R}^{D+d}$ to $\mathbb{R}^{4d}$, LSTM is implemented as:
\begin{align}
&\begin{pmatrix}\bm{i}_t\\\bm{f}_t\\\bm{o}_t\\\bm{g}_t\end{pmatrix} =
  \begin{pmatrix}\sigma\\\sigma\\\sigma\\\tanh\end{pmatrix}
  \bm{T}_{D+d,4d}\begin{pmatrix}\bm{x}_t \\ \bm{h}_{t-1}\end{pmatrix}\\
&\bm{c}_t = \bm{f} \odot \bm{c}_{t-1} + \bm{i}_t \odot \bm{g}_t\\
&\bm{h}_t = \bm{o} \odot \tanh(\bm{c}_t),
\end{align}
where $\bm{i}_t, \bm{f}_t, \bm{c}_t, \bm{o}_t$ are the input, forget, memory, output gates respectively, $\sigma$ and $\tanh$ are element-wise activation functions, $\bm{x}_t$ is the input to the LSTM in $t$-th step and $\bm{h}_t$ is the hidden state of the LSTM in the $t$-th step. 

For the simplicity of notation, we denote one computation step of the LSTM recurrence as a function \verb|LSTM|, defined as:

\begin{equation}
\bm{h}_t = \verb|LSTM|(\bm{x}_t, \bm{h}_{t-1}).
\end{equation}

\subsubsection{Reading from the Memory} \label{sec:read_op}
In this section, we describe the mechanism to read out information from the memory.
Given an external memory with buffer size $N_1$, $\bm{\mathcal{M}} = \{(\bm{m}_\text{key}^1, \bm{m}_\text{val}^1), (\bm{m}_\text{key}^2, \bm{m}_\text{val}^2), \ldots, (\bm{m}_\text{key}^{N_1}, \bm{m}_\text{val}^{N_1})\}$, where each memory slot $\bm{m}^i$ is encoded as key-value structure, \ie, $(\bm{m}^i_\text{key}, \bm{m}^i_\text{val})$, or equivalently $\bm{m}^i_\text{key}: \bm{m}^i_\text{val}$.    $\bm{m}^i_\text{key} \in \mathbb{R}^{d_1} , \bm{m}^i_\text{val} \in \mathbb{R}^{d_2}$, where $d_1$ is the dimension of the image embedding (\ie, the \verb|key| part) in the memory slot, and $d_2$ denotes the dimension of the label embedding (\ie, the \verb|val| part) in the memory slot. We use the tuple notation $(\bm{m}_\text{key}^i, \bm{m}_\text{val}^i)$ in the rest. We apply the reading mechanisms from the set-to-set framework~\cite{set2set} on the memory bank. For each time step $t$, we have:
\begin{eqnarray}
\bm{q}_t &=& \verb|LSTM|(\bm{0}, \bm{q}^*_{t-1})  \label{eq:read_LSTM} \\
e_{i,t} & = & \bm{q}_t^\top \bm{m}^i_{\text{key}}  \label{eq:q_m_dot_prod} \\
a_{i,t} & = & \verb|Softmax| (e_{i,t}) \label{eq:softmax_out}\\
\bm{z}^t_\text{key} & = & \sum_i a_{i,t} \bm{m}^i_\text{key} \label{eq:readout_key} \\
\bm{z}^t_\text{val} & = & \sum_i a_{i,t} \bm{m}^i_\text{val} \label{eq:readout_val} \\
\bm{q}^*_t &=& [\bm{q}_t,  \bm{z}^t_\text{key}].
\end{eqnarray}

\noindent $(\bm{m}^i_\text{key}, \bm{m}^i_\text{val})$, $i = 1, 2, \ldots, N_1$, are all of the memory slots stored in $\bm{\mathcal{M}}$. When the query $\bm{q}_t$ comes, it conducts dot product with all of the \verb|key| part of the memory slot $\bm{m}_\text{key}^i$ (Eqn.~(\ref{eq:q_m_dot_prod})), to obtain the similarity metric $e_{i,t}$ between query image $\bm{q}_t$ and image in the memory slot $\bm{m}_\text{key}^i$. The \verb|Softmax| operation of Eqn.~(\ref{eq:softmax_out}) generates an attention weight $a_{i,t}$ over the whole memory $\bm{\mathcal{M}}$. Then, Eqn.~(\ref{eq:readout_key}) and Eqn.~(\ref{eq:readout_val}) utilize the learned attention weight $a_{i,t}$ to read out the \verb|key| part and the \verb|value| part, \ie, label embedding, from the external memory. The readout operation blended all of the key/value vectors $\bm{m}_\text{key}^i$/$\bm{m}_\text{val}^i$ with the attention weight $a_{i,t}$ to obtain 
the readout vectors $\bm{z}^t_\text{key}$ and $\bm{z}^t_\text{val}$. Finally, $\bm{z}^t_\text{key}$ is concatenated with query $\bm{q}_t$, producing $\bm{q}_t^*$ to be fed into the next step as input of LSTM (Eqn.~(\ref{eq:read_LSTM})). The above reading procedure would loop over the memory for $T$ timesteps, obtaining $T$ readout pairs of vectors, \ie, $\{(\bm{z}^1_\text{key}, \bm{z}^1_\text{val}), (\bm{z}^2_\text{key}, \bm{z}^2_\text{val}), \ldots, (\bm{z}^T_\text{key}, \bm{z}^T_\text{val})\}$. The LSTM controller takes no input but computes recurrent state to control the reading operation. For more details, please refer the vector version (the memory slot is in the form of vector instead of key-value) of this reading mechanism~\cite{set2set}.

After $T$-step \verb|READ| operations over the memory $\bm{\mathcal{M}}$ (could be either $\bm{\mathcal{M}}_\text{ext}$ or $\bm{\mathcal{M}}_\text{abs}$), we can obtain:
\begin{equation}
\bm{\mathcal{Z}} = \{ (\bm{z}^1_\text{key}, \bm{z}^1_\text{val}), (\bm{z}^2_\text{key}, \bm{z}^2_\text{val}), \ldots, (\bm{z}^T_\text{key},  \bm{z}^T_\text{val})   \}.
\end{equation}

\subsubsection{Abstraction Memory} \label{sec:abs_mem}

We propose to utilize a novel memory component, namely \emph{abstraction memory}, into our implementation of Memory Networks. 
The abstraction memory has the following properties: 
\begin{enumerate}
\item Learn task-specific representation for the few-shot object recognition task;
\item Try to tackle the problem of efficiency of content-based addressing over a large external memory pool.
\end{enumerate}

Abstraction memory is a \emph{writable} memory bank $\bm{\mathcal{M}_\text{abs}}$, with buffer size $N_2$. It satisfies $N_2 < N_1$, where $N_1$ is the buffer size of the external memory bank $\bm{\mathcal{M}}_\text{ext}$. We denote $\bm{\mathcal{M}_\text{abs}} = \{ (\tilde{\bm{m}}^1_\text{key}, \tilde{\bm{m}}^1_\text{val}), (\tilde{\bm{m}}^2_\text{key}, \tilde{\bm{m}}^2_\text{val}), \ldots, (\tilde{\bm{m}}^{N_2}_\text{key}, \tilde{\bm{m}}_\text{val}^{N_2})  \}$, where $\tilde{\bm{m}}^i_\text{key} \in \mathbb{R}^{\tilde{d}_1}$, $\tilde{\bm{m}}^i_\text{val} \in \mathbb{R}^{\tilde{d}_2}$, $\tilde{d}_1$ is the dimension of the \verb|key| vector stored in the memory slot, and $\tilde{d}_2$ is the dimension of the \verb|value| part stored in the memory slot.

\noindent \textbf{Writing}.
Different from the \emph{external} memory bank, the \emph{abstraction} memory bank is ``writable'', which means the neural networks can learn to update the memory slots in the storage, by remembering and abstracting what matters for the specific tasks. The memory update is according to an embedding (\ie, through an \verb|FC| layer) of the readout $(\bm{z}_\text{key}, \bm{z}_\text{val})$ from the larger external memory bank $\bm{\mathcal{M}}_\text{ext}$.

Following the writing operation proposed in Neural Turing Machines (NTMs)~\cite{NTM}, we conduct the differentiable \verb|WRITE| operation on the abstraction memory bank $\bm{\mathcal{M}}_\text{abs}$.
The LSTM controller produces \emph{erase} vectors $\bm{e}_\text{key} \in \mathbb{R}^{\tilde{d}_1}$, $\bm{e}_\text{val} \in \mathbb{R}^{\tilde{d}_2}$, and \emph{add} vectors $\bm{a}_\text{key} \in \mathbb{R}^{\tilde{d}_1}$, $\bm{a}_\text{val} \in \mathbb{R}^{\tilde{d}_2}$. Note that each element of the erase vector satisfies $0 < \bm{e}^i_\text{key} < 1$ and $0 < \bm{e}^i_\text{val} < 1$, where can be implemented by passing through a Sigmoid function $\sigma(x)$.

  For each memory slot $\tilde{\bm{m}}^i$, the \verb|WRITE| operation conducts the following updates in the abstraction memory bank $\bm{\mathcal{M}}_\text{abs}$. For each timestep $t$, we have
\begin{eqnarray}
    \tilde{\bm{m}}^i_\text{key}  & \leftarrow & \tilde{\bm{m}}^i_\text{key}(\bm{1} - w_{i,t} \bm{e}_\text{key}) + w_{i,t} \bm{a}_\text{key}, \\
        \tilde{\bm{m}}^i_\text{val}  & \leftarrow & \tilde{\bm{m}}^i_\text{val}~(\bm{1} - w_{i,t}  \bm{e}_\text{val}) + w_{i, t}  \bm{a}_\text{val}. 
\end{eqnarray}

The vector $\bm{w}_t$ is used for addressing mechanisms in \verb|WRITE| operation~\cite{NTM}. However, different from NTMs, we do not utilize the location-based addressing but only the content-based addressing over the abstraction memory $\bm{\mathcal{M}}_\text{abs}$. The vector $\bm{w}_t$ can be calculated as in Eqn.~(\ref{eq:q_m_dot_prod}) and Eqn.~(\ref{eq:softmax_out}), by replacing $\bm{m}_\text{key}$ of $\bm{\mathcal{M}}_\text{ext}$ into $\tilde{\bm{m}}_\text{key}$ of $\bm{\mathcal{M}}_\text{abs}$. 

\noindent\textbf{Discussion on related works.}
Existing Memory Network variants usually only store very recent sentences in the external memory. For example, the capacity of the memory in MemN2N is restricted to the most recent 50 sentences. Douge~\etal~\cite{dodge2015evaluating} proposed to utilize inverted index to retrieve relevant sentences from a large pool to achieve efficient accessing over the external memory. KV-MemNNs utilize the same way as Douge~\etal~\cite{dodge2015evaluating} on the key hashing stage. However, inverted index is not applicable on visual ConvNet features. Our proposed abstraction memory has the same motivation to accelerate the addressing on a very large external memory pool. We address this issue in a different view,  enabling stochastic training of memory by incorporating a new writable memory component. The retrieval from a large pool of external memory is implemented in a learnable way, instead of ``hand-crafted'' inverted index as in Douge~\etal~\cite{dodge2015evaluating}.

Set-to-set (set2set) framework~\cite{set2set} also can be seen as a neural model which reads from one memory bank and writes into another memory bank. However, in the write procedure of the set2set framework, a Pointer Network (PtrNet)~\cite{vinyals2015pointer} is employed, which only allows the direct transport from the read memory to the write memory. Distinct from the set2set framework, our model enables a task-specific representation learning in the writable memory (\ie, abstraction memory in our model).

\subsubsection{Prediction} \label{sec:cls}
When it goes to the prediction stage, our model reads $(\bm{u}_\text{key}, \bm{u}_\text{val})$ from the abstraction memory $\mathcal{M}_\text{abs}$, as shown in Eqn.~(\ref{eq:read_abs}). The reading mechanism has been illustrated in Section~\ref{sec:read_op}. Reading from the memory is a recurrent process, with $T$ timesteps, we can fetch readouts $\bm{\mathcal{U}} = \{ [\bm{u}^1_\text{key}, \bm{u}^1_\text{val}], [\bm{u}^2_\text{key}, \bm{u}^2_\text{val}], \ldots, [\bm{u}^T_\text{key}, \bm{u}^T_\text{val}]\} $ to obtain enough information for few-shot classification, where $[\bm{u}^i_\text{key}, \bm{u}^i_\text{val}]$ denotes the concatenation of two vectors into one. We then run an LSTM on top of the sequence $\bm{\mathcal{U}}$, obtain the final state output $\bm{h}_T$ from the LSTM, and then feed $\bm{h}_T$ into an \verb|FC| layer and a \verb|Softmax| layer to output the prediction $\hat{\bm{y}}$. 

In this way, our model can fully utilize the readout vectors, with both visual information and label embedding information to conduct classification. These readout vectors are from abstraction memory, in which it learns to adapt in specific tasks, \eg, few-shot object recognition.
\subsection{Training}
We apply a standard cross entropy loss between the prediction $\hat{\bm{y}}$ and the groundtruth $\bm{y}$, where $\bm{y}$ is the one-hot representation of the groundtruth label.

All of the operations and components in our model are fully differentiable, which means we can train our model with stochastic gradient descent (SGD) in an end-to-end way.
\subsection{Inference}
In the inference (testing) stage, we do not make the external memory $\bm{\mathcal{M}}_\text{ext}$ available, since the abstraction memory $\bm{\mathcal{M}}_\text{abs}$ has stored all of the required information in the form of key-value in the memory slots. Thus, on the inference stage, we only run the prediction process (\cf Section~\ref{sec:cls}) on the fetched vectors from $\bm{\mathcal{M}}_\text{abs}$. The predicted label is obtained by an \verb|argmax| operation over the softmax probability output $\hat{\bm{y}}$.

%% file: expr.tex

\section{Experiments}
We evaluate our proposed model using two different external image sources, \ie, ImageNet~\cite{ImageNet} dataset and OpenImages~\cite{openimages} dataset.
In this section, we describe the specific model configurations used in the experiments, and show the results of the few-shot recognition model trained from clean human-labeled annotations and machine-labeled annotations.
Our model is implemented using TensorFlow~\cite{TensorFlow}. We make our code and the trained models publicly available upon acceptance.

\subsection{Preprocessing} We use features from top convolutional layers as image embeddings.
In all our experiments, we use the last layer activations before the final classification from the ResNet-200~\cite{he2016identity} model pretrained on ImageNet~\cite{ImageNet}.
This single model achieved top-5 error of 5.79\% on the ILSVRC 2012 validation set.
Following the standard image preprocessing practices,
images are first resized to 256 in the short side and then the central $224\times224$ subregion is cropped, we thus obtain the image embedding with
feature dimension of 2,048.
We apply the word embedding from the state-of-the-art language modeling model~\cite{jozefowicz2016exploring} in our label to word embedding mapping.
We follow the instructions provided by the authors to extract embeddings for each word in the vocabulary, and embeddings are averaged if there are multiple words for one category.
The embedding length is of 1,024 and we thus have the embedding matrix of $|V|$ by 1,024, where $|V|$ is the size of the vocabulary $V$.
The ResNet for visual feature extraction and the label embedding matrix will \emph{not} be updated during training.

\subsection{Model Specifications}
For all the LSTM models, we use one-layer LSTM with hidden unit size of 1,024. In particular, we utilize Layer Normalization~\cite{LayerNorm} for the gates and states in the cell, and we found it crucial to train our model. Layer Normalization helps to stabilize the learning procedure in RNNs, without which we could not train the network successfully.
Dropout is used in the input and output of LSTMs and we set the Dropout probability to 0.5.
The default model parameters are described following. 
We use $N_1 = 1,000$ memory slots for the external memory bank and $N_2 = 500$ memory slots for the abstraction memory.
Both \verb|key| and \verb|value| vectors stored in the abstraction memory have the dimensionality of 512.
The controller iterates $T = 5$ times when abstracts information from the external memory banks. We use the default model parameters in all the experiments unless otherwise stated.

Our model is trained with an ADAM optimizer~\cite{kingma2014adam} with learning rate at \num{0.0001} and clip the norm of the global gradients at 10 to avoid gradient exploding problem~\cite{seq2seq}. Weights in the neural network are initialized with Glorot uniform initialization~\cite{glorot2010understanding} and weight decay of \num{0.0001} is applied for regularization.

\subsection{Datasets}
\noindent \textbf{ImageNet}: ImageNet is a widely used image classification benchmark. There are two sets in the ImageNet dataset. One part is used in the ILSVRC classification competitions, namely ILSVRC 2012 CLS. This part contains exactly 1,000 classes with about 1,200 images per class which has well-verified human-labeled annotations. The other set of ImageNet is the whole set, which consists of about 21,000 categories.

\begin{figure}[h]
\centering
\includegraphics[width=0.95\linewidth]{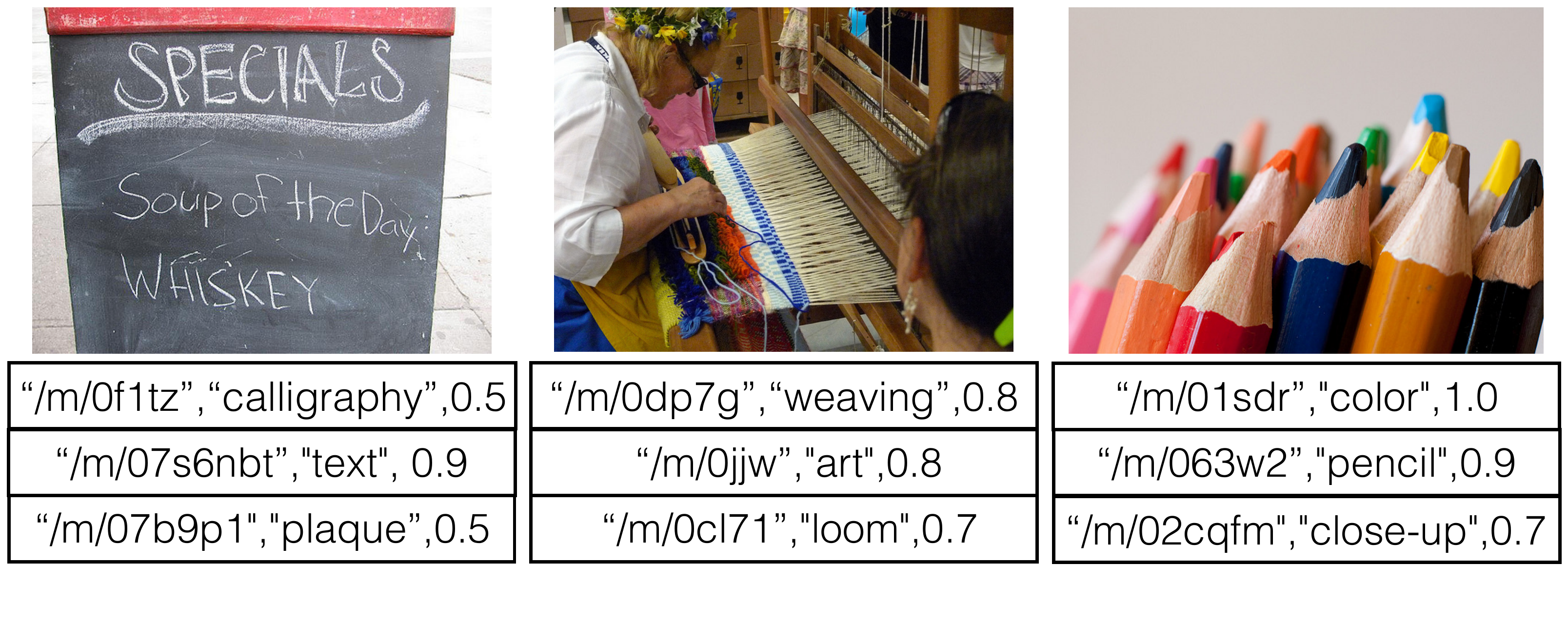}
    \caption{Sample images from the OpenImages dataset. Annotations on the images are shown in the bottom lines. The annotations listed are ``label id'', ``label name'', ``confidence'' tuples.}
\label{fig:example_openimage}
\end{figure}

\noindent \textbf{OpenImages}. The recently released OpenImages dataset~\cite{openimages} consisting of web images with machine-labeled annotations.
In the original dataset, it is split into a training set with 9,011,219 images and a validation set with 167,057 images.
There are 7,844 distinct labels in OpenImages, whose label vocabulary and diversity is much richer than the ILSVRC 2012 CLS dataset.
Since this dataset is pretty new, we provide some example images in Figure~\ref{fig:example_openimage}.
We can see that the OpenImages dataset has a wider vocabulary than ImageNet ILSVRC 2012 dataset,
which is beneficial for the generalization to novel categories.

\subsection{Few-shot Learning with Human-labeled\\annotations}
We first validate our model on the task of few-shot classification using human-labeled clean data.

For few-shot image classification, the training set has only a few examples, and the basic task can be denoted as $N$-way $k$-shot classification (following the notation of Matching Networks~\cite{vinyals2016matching}),
in which $N$ classes images need to be classified and each class is provided with $k$ labeled examples ($k$ is usually less than 10).

\noindent\textbf{Dataset}.
We now construct our dataset for few-shot learning.
We select 100 classes for learning by randomly choosing 100 categories from the whole 21,000 categories in the ImageNet dataset, excluding the 1,000 categories in the ILSVRC 2012 CLS vocabulary.
For testing, there are 200 images per category and the training set have $k$ examples per category.
We use settings of $k=1$, $k=5$, $k=10$,~\ie, there are 1 example, 5 examples and 10 examples in the training set.

\subsubsection{Comparison with other methods}
In this experiment, we use image-label pairs from the ILSVRC 2012 CLS dataset as external memory. We use all 1,000 categories for learning.
We conduct experiments on 1-shot, 5-shot, 10-shot tasks and compare with several algorithms.
The results are shown in Table~\ref{expr_clean_imagenet_100_cats}.

\noindent\textbf{$k$-NN} and \textbf{Exemplar-SVMs}. $k$-Nearest Neighbors ($k$-NN) is a simple but effective classifier when very few training examples are provided.
We utilize ResNet-200 features and consider two distance metrics, \ie, $l_1$ and $l_2$, for pairwise distance calculation.
Exemplar-SVMs (E-SVM)~\cite{exemplar} train an SVM for each positive example and ensemble them to obtain the final score. The method had been widely used in object detection in the pre-ConvNet era. We use the same ResNet-200 features and set $C=0.1$.
The results show that our methods outperforms $k$-NN for both $l_1$ and $l_2$ distance with a large margin and it also outperforms E-SVMs.
Note that on 5-shot and 10-shot tasks, our model achieves better performance than the E-SVMs with larger margin.
The results show that our model can take advantage of the large number of image-label pairs in the external memory by learning relationships
between the examples and the external data.

\noindent\textbf{KV-MemNNs}.
By utilizing the interpretation of image embedding as \verb|key| and label embedding as \verb|value| as in our model, KV-MemNNs can also be trained to conduct few-shot learning.
However, due to the design of KV-MemNNs, the few-shot prediction has to rely on the external memory, while the utilized image classification datasets in our work are too large to be stored in.
This property makes KV-MemNNs has non-deterministic classification prediction, which is not preferable.
It is unrealistic to search over all image-pairs in the external memory during each training iteration.
In the testing, it is also time-consuming to traverse the whole external memory.
As a workaround, we randomly sample 1,000 pairs from the external memory for matching during both training and testing.
To alleviate the randomness results in the testing, we report the mean classification results and the standard deviation in 20 runs.
The result shows that our abstraction memory extracted valuable information from the large external memory and is much more compact
than the original memory banks.

\begin{table}[t]
\begin{center}\begin{tabular}{|l|c|c|c|}
\hline
    Methods                                                   &    1-shot  &  5-shot  &  10-shot  \\
\hline\hline
    $k$-NN ($l_1$)                                          &  38.8     &  57.0   &  62.9     \\
\hline
    $k$-NN ($l_2$)                                          &  38.6     & 56.4    &   62.1    \\
\hline
    E-SVM                                                     &  45.1     &  62.3   &   68.0   \\
\hline
    KV-MemNNs                                                     & 43.2 (\footnotesize{$\pm$0.4}) & 66.6 (\footnotesize{$\pm$0.2})   &  72.8 (\footnotesize{$\pm$0.2}) \\
\hline
    Ours                                                      &  \textbf{45.8}    &  \textbf{68.0}    &  \textbf{73.5}   \\
\hline
\end{tabular}\end{center}
\caption{Comparison between our model with other methods. Results are reported on our 100-way testing set.}
\label{expr_clean_imagenet_100_cats}
\end{table}

\noindent\textbf{Matching Networks}. We also compare to the recently proposed Matching Networks~\cite{vinyals2016matching}.
They use two embedding functions considering set context. However, as LSTMs are used for embeddings, the size
of support set is limited.
In~\cite{vinyals2016matching}, the number of categories is usually set to 5 for ImageNet experiments (5-way).
For fair comparison, we conduct experiment on the 5-way 1-shot task and the model is implemented by our own
as the authors have not released the source code.
We randomly choose 5 categories from the previous used 100 categories set. The testing set has the same number of instances per category.
The result is shown in Table~\ref{expr_cmp_with_vinyals}.
It shows that our method outperforms the Matching Network. Our model builds explicit connection between the few training examples and the external memory which could benefit a lot from large vocabulary.

\begin{figure}[h]
\centering
\includegraphics[width=0.95\linewidth]{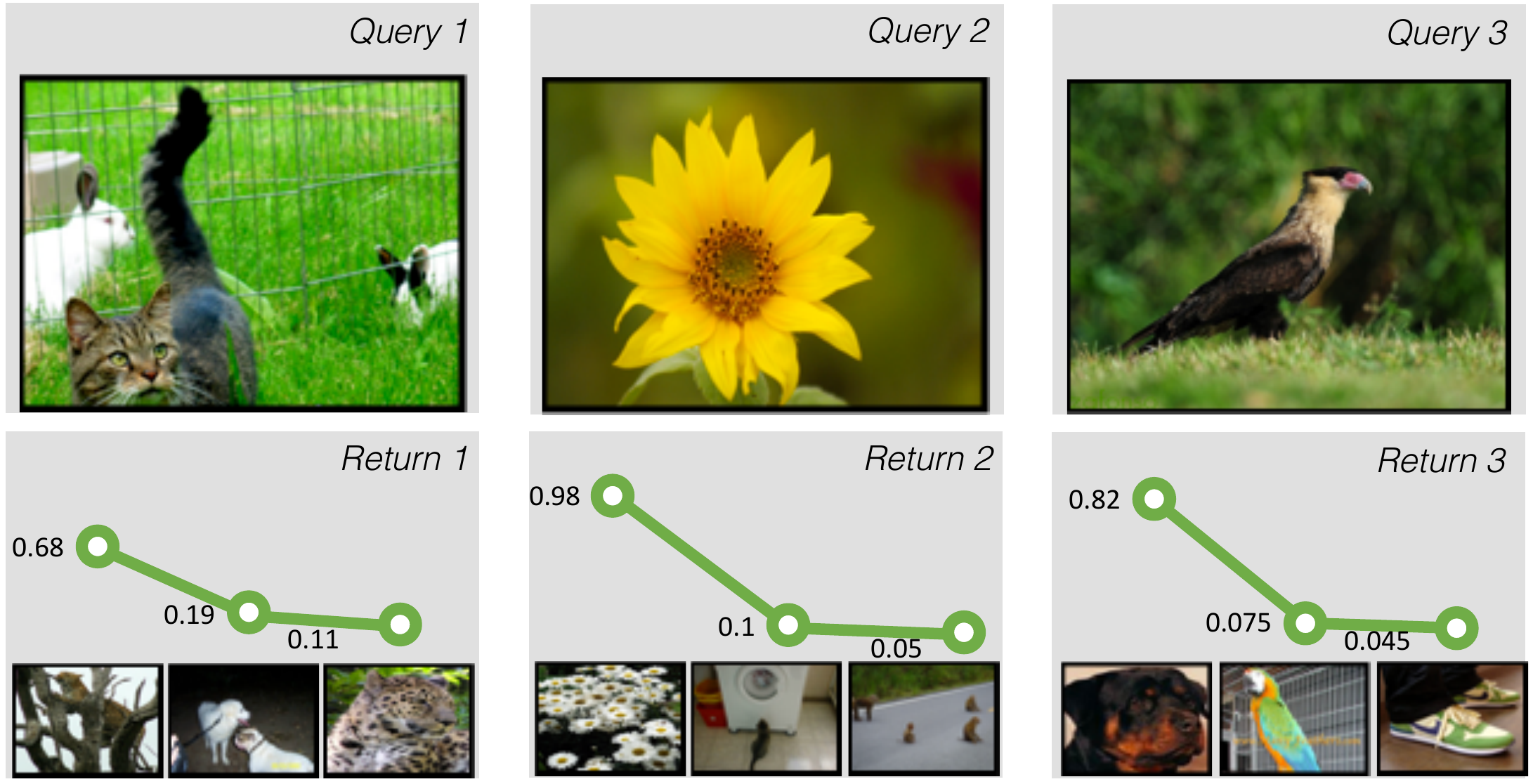}
\caption{We show the query results returns from the external memory. The scores are the softmax probabilities. Only top-3 results are shown.}
\label{fig:example_matching}
\end{figure}

We visualize the query results between the external memory and the query in Figure~\ref{fig:example_matching}.

\begin{table}[t]
\begin{center}\begin{tabular}{|l|c|c|c|}
\hline
    Methods                                      &   5-way 1-shot   classification \\
\hline\hline
    Matching Networks                            &      90.1        \\
\hline
    Ours                                         &      \textbf{93.9}       \\
\hline
\end{tabular}\end{center}
    \caption{Comparison between our model with Matching Networks on the 5-way 1-shot task.}
\label{expr_cmp_with_vinyals}
\end{table}

\subsection{Few-shot Learning with Machine-labeled\\annotations}
In this experiment, we replace the external memory source with the OpenImages dataset.
The machine-labeled images are much easier to obtain but are noisier.
We train our model to learn from such noisy web images.

We construct the external memory using the OpenImages dataset.
We use four different external memory settings which are, 1,000 vocabulary with human-labeled images, 1,000 vocabulary with machine-labeled images,
6,000 vocabulary with human-labeled images, and 6,000 vocabulary with machine-labeled images.
Note that although the OpenImages dataset is machine-labeled, the validation set in the original dataset is also validated by human raters.
The results are shown in Table~\ref{expr_openimages_vocab}.
It shows that machine-labeled external memory can serve as a good source for few-shot learning, which
is worse than the human-labeled external memory with about 1\%.

Besides, as the vocabulary size growing, we can observe that the performance improves.
It shows that with large vocabulary, our model is able to reason among the external memory in a more effective way.
Larger vocabulary will be explored in the future.

\begin{table}[t]
\begin{center}\begin{tabular}{|l|c|c|c|}
\hline
    Methods                                      &   1,000   &    6,000  \\
\hline\hline
    Machine-labeled                               &   66.6    &    67.4   \\
\hline
    Human-labeled                                &   67.7    &    68.2  \\
\hline
\end{tabular}\end{center}
\caption{Results on the OpenImages dataset. The results are reported on the 100-way 5-shot task.}
\label{expr_openimages_vocab}
\end{table}

%% file: conclusion.tex

\section{Conclusion}
We propose a novel Memory Networks architecture specifically tailored to tackle the few-shot learning problem on object recognition. By incorporating a novel memory component into the Key-Value Memory Networks, we enable the rapid learning of just seeing a handful of positive examples by abstracting and remembering the presented external memory. We utilize LSTM controllers for reading and writing operations into the memories. We demonstrated that our proposed model achieves better performance over other state-of-the-art methods. Furthermore, we obtain similar performance by utilizing machine-labeled annotations compared to human-labeled annotations. Our model paves a new way to utilize abundant web image resources effectively. We will explore the structural storage of external memory for different vision datasets, or for different tasks, \eg, image classification~\cite{ResNet}, object detection~\cite{faster_RCNN}, human pose estimation~\cite{chu2016crf}, and learn task-specific representation on demand in our abstraction memory framework.